*Article*

# Multi-view registration of unordered range scans by fast correspondence propagation of multi-scale descriptors

**Jihua Zhu** [1,*], **Siyu Xu** [1], **Zutao Jiang**[1], **Shanmin Pang**[1], **Jun Wang**[2] **and Zhongyu Li**[3]

[1] School of Software Engineering, Xi'an Jiaotong University, Xi'an 710049, China
[2] School of Digital Media, Jiangnan University, Wuxi 214122, China
[3] University of North Carolina at Charlotte, NC, USA
* Correspondence: jhzhu@mail.com; Tel.: +86-29-82663000-8051

**Abstract:** This paper proposes a global approach for the multi-view registration of unordered range scans. As the basis of multi-view registration, pair-wise registration is very pivotal. Therefore, we first select a good descriptor and accelerate its correspondence propagation for the pair-wise registration. Then, we design an effective rule to judge the reliability of pair-wise registration results. Subsequently, we propose a model augmentation method, which can utilize reliable results of pair-wise registration to augment the model shape. Finally, multi-view registration can be accomplished by operating the pair-wise registration and judgment, and model augmentation alternately. Experimental results on public available data sets show, that this approach can automatically achieve the multi-view registration of unordered range scans with good accuracy and effectiveness.

**Keywords:** Range scan; correspondence propagation; multi-view registration; pair-wise registration; model augmentation

## 1. Introduction

Due to the wide applications, range scan registration has been attracted more and more attentions in many domains, such as robot mapping [1-4], computer vision [5,6] and computer graphics [7,8]. Given the global reference frame, the goal of registration is to find optimal transformation(s) for one or more scans so as to transfer all scans into the unified frame. According to number of scans to be registered, this problem can be divided into pair-wise registration and multi-view registration.

For pair-wise registration, Besl et al. [9] proposed the iterative closest point (ICP) algorithm, which is one of the most popular registration approaches. Although this approach can achieve the pair-wise registration with good efficiency, it is unable to get desired results for registration of partially overlapping range scans. Besides, it is a local convergent approach. To obtain the desired results, good initial parameters should be provided to the ICP algorithm. For registration of partially overlapping scans, Chetverikov et al. [10] proposed the trimmed ICP (TrICP) algorithm, which introduces the overlap percentage to determine the overlapping parts so as to estimate accurate rigid transformation. Further, Phillips et al. [11] improve the performance of original TrICP algorithm in efficiency. To address local convergence, Genetic algorithm (GA) [12-14] and particle filtering [15] have been applied for global registration of scan pairs. To guarantee global optimality, initializations still require being reasonably good as otherwise spaces of parameters are too large for heuristic search. What's more, these global optimization approaches are requires high computation complexity. Subsequently, Yang et al. [16] proposed the first globally optimal algorithm, named Go-ICP. This approach is based on a branch-and-bound (BnB) scheme that searches the entire 3D motion space and able to produce reliable registration results regardless of the initialization. Besides, many feature match based approaches have been proposed for pair-wise registration [17-19]. Among these approaches, one of the most efficient and robust approaches is multi-scale descriptors with correspondence propagation [19], which can be further improved in efficiency.

Although these above-mentioned approaches may obtain good results for pair-wise registration, most of them cannot deal with multi-view registration. Given a set of range scans, the task of multi-view registration is to find optimal transformation for each scans to the reference scan. Compared with pair-wise registration, multi-view registration is somewhat difficult due to many transformations required to be estimated. To solve this problem, many good solutions have been proposed. Among these solutions, the sequential registration approach is the simplest one [20]. For multi-view registration, it alternately aligns and integrates two scans until all scans are integrated into one model. Although this approach is simple, it suffers from the error accumulation problem. To address this issue, Zhu et al. [21] proposed the coarse-to-fine registration approach, which sequentially aligns each scan to a model integrated by all other align scans. By traversing all scans, their transformations are sequentially refined by the pair-wise registration. Meanwhile, Govindu et al. [22] proposed multi-view registration approach based on motion averaging algorithm, which takes a set of available relative motions (pair-wise registration results) as input to simultaneously estimate all transformations for multi-view registration. Based on this work, Guo et al. [23] proposed the weighted motion algorithm for multi-view registration. This approach can improve the robustness and accuracy of multi-view registration by paying more attention to reliable pair-wise registration results. Recently, Arrigoni et al. [24] proposed multi-view registration approach based on the low-rank and sparse (LRS) matrix decomposition. It stacks all available relative motions into a large matrix and replaces the unavailable ones with zeros, then utilizes the LRS decomposition to recover global motions for multi-view registration. Moreover, Georgios et al. [25] cast the multi-view registration problem into the framework of clustering. For clustering, it utilizes the Expectation-Maximization (EM) algorithm to simultaneously estimate Gaussian Mixture Model (GMM) and all transformations for multi-view registration.

Although these above approaches may obtain desired results for multi-view registration, they should be provided with good initial registration parameters. Otherwise, it is difficult to obtain desired registration results. For global multi-view registration, Daniel et al. [26] proposed the approach for fully automatic registration of multiple range scans. It applied the pair-wise registration on all scan pairs involved in multi-view registration and then tested these results for surface consistency. Accordingly, reliable results of pair-wise registration can be checked for the estimation of multi-view registration. Since there are many scan pairs involved in multi-view registration, the computation complexity of this approach is very high. Subsequently, Guo et al. [27] proposed a feature matched approach for global multi-view registration. This approach is also based on the pair-wise registration and it is more efficient than previous approaches. However, it sometimes cannot find the minimum number of good feature matches for pair-wise registration, which leads to the failure of multi-view registration. Recently, Zhu et al. [28] proposed the global multi-view registration approach based on the construction of spanning tree. It designed the dual criterion to judge the reliability of pair-wise registration results. By viewing the first scan as the root node, all other scans can be added by the breadth-first search with the reliable results of the pair-wise registration. As most scan pair contains low overlap percentages, it is also required to align and check many scan pairs to construct a full spanning tree. Therefore, its computation complexity is also high.

For multi-view registration, this paper proposes a global approach for registration of unordered range scans. The contribution of this paper can be delivered as follows: 1) Improve the global pair-wise registration approach based on fast correspondence propagation of multi-scale descriptors; 2) Design the effective rule to judge the reliability of pair-wise registration; 3) Propose good model augmentation, which can properly augment the model shape for further pair-wise registration. To demonstrate its effectiveness, the proposed approach will be tested on public available data sets and compared with two related registration approaches.

The reminder of this paper is organized as follows. We briefly review pair-wise registration approach in Section 2. In Section 3, we present details of the proposed multi-view registration approach. All experimental results are displayed in Section 4. Finally, we conclude this work in Section 5.

## 2. Pair-wise registration

This section briefly reviews the TrICP algorithm and correspondence propagation of multi-scale descriptors for pair-wise registration.

*2.1. TrICP algorithm*

Suppose there are two partially overlapping scans in $\mathbb{R}^3$, a data shape $P \triangleq \{\mathbf{p}_i\}_{i=1}^{N_p}$ and a model shape $Q \triangleq \{\mathbf{q}_j\}_{j=1}^{N_q}$ ($N_p, N_q \in \mathbb{N}$), where $\xi$ denotes the overlap percentage and $P_\xi$ indicates the overlapping part of $P$ to $Q$. Therefore, the overlap percentage can be calculated as $\xi = N_p' / N_p$, where $N_p'$ denotes the number of point included in the subset $P_\xi$. Subsequently, it is convenient to define trimmed mean square error (TMSE) as:

$$e(\xi, \mathbf{R}, \vec{t}) = \frac{1}{N_p'} \sum_{\vec{p}_i \in P_\xi} (\|\mathbf{R}\vec{p}_i + \vec{t} - \vec{q}_{c(i)}\|_2^2) \qquad (1)$$

where $\mathbf{R}$ is the $3 \times 3$ rotation matrix, $\vec{t}$ is the $3 \times 1$ translation vector and $(\mathbf{p}_i, \mathbf{q}_{c(i)})$ denotes a pair of point correspondence. According to [], the task of pair-wise registration is to find an optimal rigid transformation $(\mathbf{R}, \vec{t})$, which can be obtained by minimizing the following objective function:

$$\psi(\xi, \mathbf{R}, \vec{t}) = \frac{e(\xi, \mathbf{R}, t)}{(\xi)^{1+\lambda}} \qquad (2)$$

where $\lambda (\lambda = 2)$ is a preset parameter.

Actually, Eq. (1) has been solved by the TrICP algorithm, which achieves pair-wise registration by iterations. Given initial parameters $(\mathbf{R}_0, \vec{t}_0)$, three steps are included in each iteration:

(1) Establish point correspondences between two range scans:

$$c_k(i) = \arg\min_{j \in \{1,2,\dots,N_q\}} \|\mathbf{R}_{k-1}\vec{p}_i + \vec{t}_{k-1} - \vec{q}_j\|_2, i=1,2,\dots,N_p \qquad (3)$$

(2) Update the overlap percentage and its corresponding subset:

$$(\xi_k, P_{\xi_k}) = \arg\min_{\xi_{\min} < \xi \leq 1} \sum_{\vec{p}_i \in P_\xi} \|\mathbf{R}_{k-1}\vec{p}_i + \vec{t}_{k-1} - \vec{q}_{c_k(i)}\|_2^2 / (|P_\xi| \xi^{1+\lambda}) \qquad (4)$$

(3) Calculate the new transformation:

$$(\mathbf{R}_k, \vec{t}_k) = \arg\min_{\mathbf{R}, \vec{t}} \sum_{\vec{p}_i \in P_{\xi_k}} \|\mathbf{R}\vec{p}_i + \vec{t} - \vec{q}_{c_k(i)}\|_2^2 \qquad (5)$$

A rigid transformation $(\mathbf{R}, \vec{t})$ will be obtained by repeating step (1)-(3) until the iteration $k$ reaches the maximum number $K$ or $|\psi_k - \psi_{k-1}| < \varepsilon$, where $\varepsilon$ denotes a given small positive number. Similar to the original ICP algorithm, the TrICP algorithm is also local convergent. To obtain desired results, good initial parameters should be provided.

*2.2 Correspondence propagation of multiscale descriptors*

For the pair-wise registration, multi-scale descriptors have been proposed in []. Given a point, $L$ different support radii are considered. Within each support radius, all included range points are used to compute one variance matrix $\mathbf{C}_l$. Based on these $L$ circles and variance matrix, it is convenient to compute multi-scale descriptors $(\mathbf{N}, \mathbf{D})$, where $\mathbf{N}$ denotes the normal vector and $\mathbf{D}$ indicates the eigenvalue based vector. Accordingly, the pair-wise registration is solved by the effective approach displayed in Fig. 1.

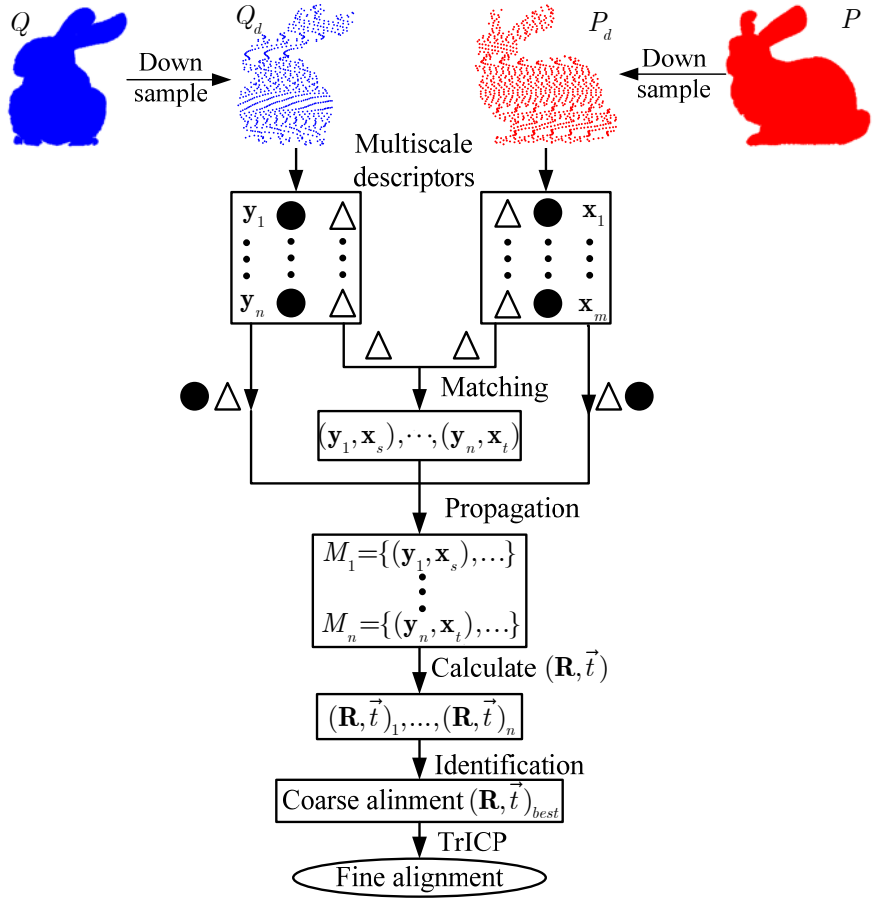

Fig. 1 Flow chart of the pair-wise registration approach based on correspondence propagation of multi-scale descriptors

As shown in Fig. 1, after the calculation of multi-scale descriptors, the correspondence of each point from $y_1$ to $y_n$ is established by performing the nearest neighbor (NN) search on the eigenvalue based vector $D$. For correspondence propagation, each of these point correspondence can be viewed as a seed match. Then, the propagation session exploits on both $D$ and $N$ to propagate each seed match (e.g., $(y_1, x_s)$ ) into a set of matches (e.g., $M_1 = \{(y_1, x_s),...\}$). For each match set, the RANSAC algorithm should be used to find the maximum consensus set so as to calculate one optimal rigid transformation. Based on down sampled scans $Q_d$ and $P_d$, a quality function formulated from distances errors is defined to identify the best one from all obtained rigid transformations. Finally, the TrICP algorithm refines results of pair-wise registration by taking the best transformation $(R,t)_{best}$ as input.

According to [19], the time complexity of each propagation operation is about $O((m+n)\log n)$, where $n$ and $m$ denotes the number of points included in the down sampled scans $Q_d$ and $P_d$, respectively. As the required number of propagation is $n$, the total complexity of correspondence propagation is $O(n(m+n)\log n)$. Although this approach is robust for the global pair-wise registration, the correspondence propagation is too time-consuming.

## 3. Global registration of multi-view scans

For multi-view registration of unordered scans, a global approach is proposed as shown in Fig. 2. Given a set of unordered scans, the frame of reference, without loss of generality, can be attached, to the first scan. As shown in Fig. 2, it achieves multi-view registration by pair-wise registration, reliability judgment of pair-wise registration and model augmentation, which are main operations of the proposed approach. Therefore, we will present more details of these three operations and the implementation of this approach.

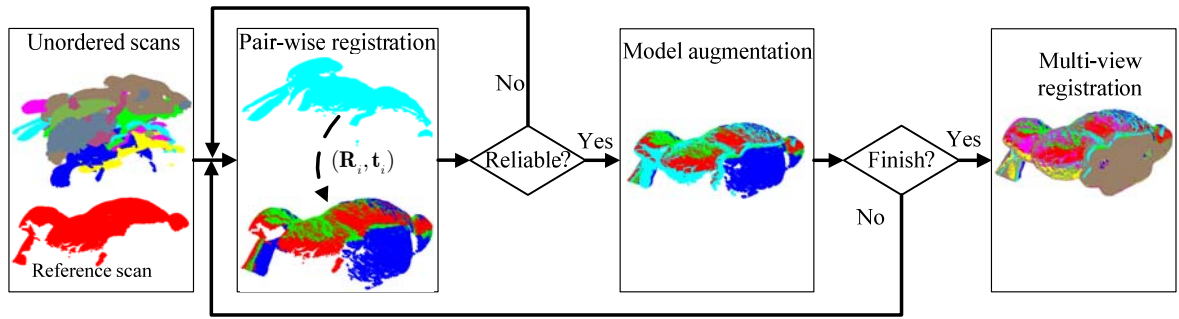

Fig. 2 The flowchart of the proposed approach for multi-view registration, which operates pair-wise registration, the judgment of pair-wise registration result and model augmentation alternately until all range scans are integrated into one model.

*3.1. Pair-wise registration by fast correspondence propagation*

In global registration, these are no prior information about initial registration parameters. Therefore, Lei et al. [19] proposed the multi-scale descriptors with correspondence propagation for global pair-wise registration. As shown in Fig. 1, this approach should do several operations to achieve pair-wise registration. Among these operations, correspondence propagation is the most important and time-consuming one.

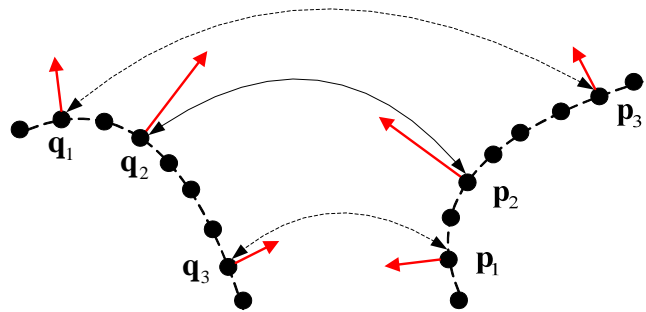

Fig. 3 Demonstration of different seed matches, one solid line with bi-directional arrow denotes a reliable seed match, each dotted line with bi-directional arrow denotes an unreliable seed match, the direction and length of red line with arrow denote the normal vector and the eigenvalue based vector, respectively.

According to [19], multi-scale descriptors consist of the normal vector $\mathbf{N}$ and the eigenvalue based vector $\mathbf{D}$, where $\mathbf{D}$ is rotation invariant. Hence, seed matches is established by performing the NN search on $\mathbf{D}$. But these seed matches may be unreliable due to the low dimension of $\mathbf{D}$. To obtain good point matches, the correspondence propagation is necessary to be applied. However, we think it is not required to apply correspondence propagation on each seed match. Because once the propagation is starts from one reliable seed match, the exploitation of $\mathbf{N}$ and $\mathbf{D}$ will certainly propagate it into a set of good matches. Hence, some reliable seed matches are enough for the correspondence propagation of multi-scale descriptors. The problem is how to find reliable seed matches. Fig. 3 illustrates both reliable and unreliable seed matches. Actually, $(\mathbf{p}_i, \mathbf{q}_i)_{i=1}^3$ denote three reliable matches. But $(\mathbf{p}_1, \mathbf{q}_3)$ and $(\mathbf{p}_3, \mathbf{q}_1)$ may be mismatched due to noises or other reasons. Therefore, one seed match $(\mathbf{p}_1, \mathbf{q}_3)$ or $(\mathbf{p}_3, \mathbf{q}_1)$ with small distance may not be reliable. While, the reliable seed match $(\mathbf{p}_2, \mathbf{q}_2)$ must contain small distance. According to this fact, it is better to sort all established seed matches in ascending order by their distances, then the correspondence propagation is only applied to top $\delta \times 100\%$ sorted seed matches. In this case, the total time complexity of correspondence propagation can be seriously reduced so as to result in efficient pair-wise registration. Based on what discussed above, an efficient pair-wise registration algorithm is summarized in Algorithm 1.

---

**Algorithm 1** Efficient pair-wise registration algorithm

**Input**: Scan pair $P$ and $Q$

  Get $P_d$ and $Q_d$ by down sample;

  Calculate descriptors $(\mathbf{N}, \mathbf{D})$ for each point in $P_d$ and $Q_d$;

  Establish seed matches by performing the NN search on $\mathbf{D}$;

  Sort all seed matches in the ascending order by their distances of $\mathbf{D}$;

  For $i = 1 : \delta n$

    Propagate the $i$ th seed match into a set of matches $M_i = \{(\mathbf{y}_i, \mathbf{x}_{c(i)}), ...\}$;

    Find maximum consensus set in $M_i$ by RANSAC;

    **If** ($|M_i| > 10$)

      Use $M_i$ to calculate $(\mathbf{R}, \mathbf{t})_i$;

    **End**

  **End**

  Find the best alignment $(\mathbf{R}, \mathbf{t})_{best}$ by the quality function;

  Utilize TrICP to refine $(\mathbf{R}, \mathbf{t})_{best}$ and obtain $(\mathbf{R}, \mathbf{t})$.

**Output:** Pair-wise registration results $(\mathbf{R}, \mathbf{t})$

---

In this algorithm, the required number of correspondence propagation is $\delta n$. As the propagation of each seed match is same to that of the method proposed in [19], the total complexity of correspondence propagation is reasonably reduced to $O(\delta n(m+n)\log n)$. Obviously, small value of $\delta$ will lead to low time complexity. Since the seed match with small distance may not be reliable, the value of $\delta$ should not be set to small. Otherwise, there may be no way to apply correspondence propagation to reliable seed match, which will lead to unexpected pair-wise registration. In practice, $\delta = 0.3$ can obtain satisfactory registration results.

*3.2. Reliability judgment of pair-wise registration*

For these scan pairs involved in multi-view registration, some of them may contain high overlap percentages, other may have lows or even no overlap percentage. Given a pair of scans, one rigid transformation will be estimated by the proposed pair-wise registration algorithm. For the scan pair with high overlap percentage, the estimated transformation is reliable. However, it is unreliable for these scan pairs without high overlap percentages. Usually, it is difficult to predict or estimate the overlap percentage of each scan pair, so the pair-wise registration may be applied to scan pair without high overlap percentage and obtain unreliable transformation. As shown in Fig. 1, only reliable results can be utilized for the model augmentation. Therefore, it is required to judge the reliability of pair-wise registration.

As the TrICP algorithm is finally utilized to refine the result of pair-wise registration, it also determines overlapping areas and provides TMSE for each scan pair. Fig. 2 displays pair-wise registration results of two scan pairs, where one scan pair gets reliable registration due to its high overlap percentage and the other one gets unreliable registration due to its low overlap percentage or other reasons. As shown in Fig. 2, the TrICP algorithm determines suitable overlapping areas with small TMSE for reliable registration and finds false overlapping areas with very large TMSE for the unreliable registration. Therefore, TMSE is related to the reliability of pair-wise registration.

For the same registration result, the TMSE is affected by the point resolution of model shape. Fig. 3 demonstrates the same pair-wise registrations under different point resolutions of model shape. As shown in Fig. 3, high resolution reduces the TMSE and low resolution increases the TMSE. Therefore, we can judge the pair-wise registration is reliable if the TMSE satisfies the following condition:

$$TMSE \leq 2d_Q \tag{6}$$

where $d_Q$ denotes the point resolution of model shape. Otherwise, the pair-wise registration is judged to be unreliable. To achieve multi-view registration, the pair-wise registration should be implemented many times in the proposed approach. During multi-view registration, it is easy to collect TMSEs from reliable registrations, which can be further utilized to judge the reliability of posterior registration results. Therefore, it is better to revise Eq. (6) as follows:

$$TMSE \leq \max(2d_Q, 1.5 m_{TMSE}) \tag{7}$$

where $m_{TMSE}$ indicates the mean TMSE of reliable registrations. Before the first reliable registration has been checked, $m_{TMSE}$ is set as $m_{TMSE} = 0$. After the check of one new reliable registration, it can be further updated.

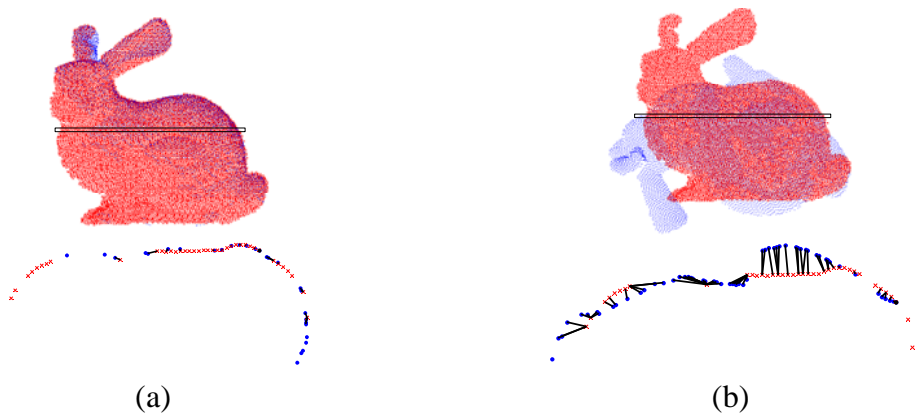

(a) (b)

Fig. 4 The illustration of TMSE for pair-wise registration of two scan pairs. In the 2nd row, each black line denotes the distance of one point pair located in the overlapping area. (a) Reliable registration. (b) Unreliable registration.

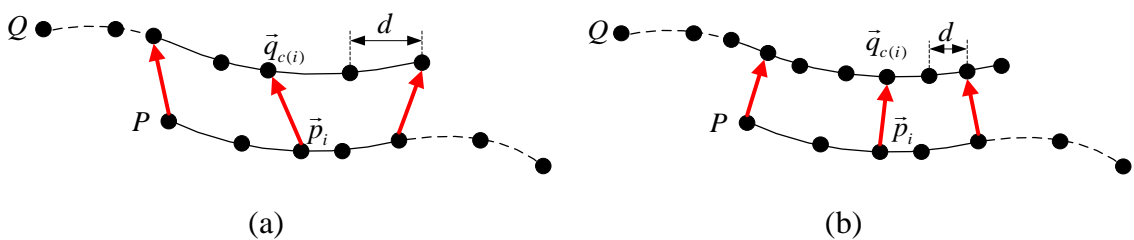

(a) (b)

Fig. 5 TMSEs under different point resolutions of model shape. (a) Low resolution results in large trimmed MSE. (b) High resolution reduces the trimmed MSE.

*3.3. Model augamentation*

By the proposed judgment, it is easy to judge whether the pair-wise registration is reliable or not. For reliable registration, data shape should be transformed and merged into the model shape. The simplest method is to directly add points of transformed data shape into the model shape. However, this rude augmentation will result in redundant range point with multi-scale descriptors, which can reduce the efficiency of subsequent registration. Hence, model augmentation should be carefully designed. Accordingly, we propose the model augmentation illustrated in Fig. 4.

Based on the estimated rigid transformation $(\mathbf{R}, \vec{t})$, the data shape is first transformed into the reference frame of model shape as follows:

$$P' \triangleq \{\mathbf{R}\mathbf{p}_i + \mathbf{t}\}_{i=1}^{N_p}. \tag{8}$$

Then both model shape and transformed data shape are divided into two independent parts:

$$Q = Q_\xi \cup A, \quad P' = P'_\xi \cup B, \tag{9}$$

where $Q_\xi$ and $P_\xi$ denotes overlapping parts of these two shapes, $A$ and $B$ indicates non-overlapping parts of them. Accordingly, the augmented model shape is composed of raw model shape and transformed data shape as follows:

$$Q' = A \cup F \cup B \tag{10}$$

where $F = \{\mathbf{f}_i\}_{i=1}^{N_f}$, $N_f = \xi N_p$, and

$$\mathbf{f}_i = \frac{(\mathbf{R}\mathbf{p}_i + \mathbf{t}) + \mathbf{q}_{c(i)}}{2}. \tag{11}$$

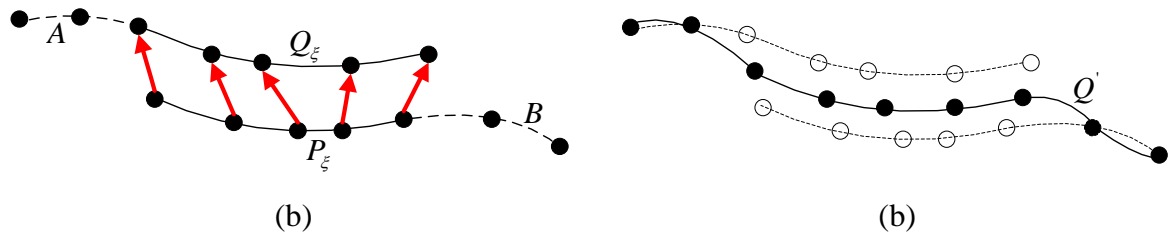

(b)　　　　　　　　　　　　　　(b)

Fig. 6 Illustration of model augmentation. (a) The point pairs located in the overlapping area, where each pair of points is connected by one line with arrow. (2) Model and data shape is merged into one shape, where each point pair is replaced by its intermediate point in the overlapping area.

Similar to these two raw shapes, the down sampled data shape $P_d$ should be correspondingly transformed and merged into the down sampled model shape $Q_d$. And they can be merged in the same way as these raw shapes. Besides, multi-scale descriptors should also be merged. As eigenvalue based vectors $\mathbf{D}$ are rotation-invariant, they can be merged as follows:

$$\mathbf{D}_{Q',i} = \frac{(\mathbf{D}_{P,i} + \mathbf{D}_{Q,c(i)})}{2}, \tag{12}$$

where $\mathbf{D}_{P,i}$ and $\mathbf{D}_{Q,c(i)}$ denote eigenvalue based vectors of two matched points. While, normal vectors $\mathbf{N}$ are rotation-invariant, so they can be merged as follows:

$$N_{Q',i} = \frac{\mathbf{R}N_{P,i} + N_{Q,c(i)}}{2} \tag{13}$$

where $\mathbf{N}_{P,i}$ and $\mathbf{N}_{Q,c(i)}$ indicate normal vectors of two matched points. In this way, the model shape is properly augmented with corresponding multi-scale descriptors, so there is no need to recalculated multi-scale descriptors for the augmented model shape.

Compared with rude augmentation, the proposed augmentation method can delete redundant range point with multi-scale descriptors.

*3.4. Implementation*

The overall process of the proposed approach can be summarized in **Algorithm** 2.

---

**Algorithm 2**：Multi-view registration of unordered range scans

**Input**: A scan set with $N$ unordered scans $\{P_i\}_{i=1}^{N}$
  View the 1st scan as $Q$ and delete it from scan set;
  Set $mMSE = 0$, $N' = 0$
  **Do**
    $N' = 0$;
    **For** i= 1: $N$
      View $P_i$ as $P$, estimate $(\mathbf{R}_i, \mathbf{t}_i)$ and $TMSE_i$ by **Algorithm 1**；
      Use Eq. (7) to judge the reliability of $(\mathbf{R}_i, \mathbf{t}_i)$;
      **If** ($(\mathbf{R}_i, \mathbf{t}_i)$ is reliable)
        $N' = N' + 1$;
        Update $Q$ by the **Sec. 3.3**；
        Delete $P_i$ from scan set；
        Recalculate $m_{TMSE}$ and $d_Q$
      **End**
    **End**
    $N = (N - N')$
  **While** ($N > 0$)
**Output**：Multi-view registration results $\{(\mathbf{R}_i, \mathbf{t}_i)\}_{i=1}^{N}$

---

Given a set of unordered range scans, the proposed approach can automatically achieve global multi-view registration without any prior information.

**4. Results and Discussion**

To illustrate its performance, experiments were conducted on four data sets taken from the Stanford 3D Scanning Repository [29], which provide multi-view scans with ground truth of rigid transformations $\{(\mathbf{R}_{i,g}, \mathbf{t}_{i,g})\}_{i=1}^{N}$. In the calculation of multi-scale descriptors, raw range points of each scan were down sampled with the sampling frequency set to be 100. Besides, raw range points of each scan were down sampled with the sampling frequency set to be 10 for the pair-wise registration. For facilitate comparison, rotation error and translation error are defined as $e_{\mathbf{R}} = \frac{1}{N}\sum_{i=1}^{N}\|\mathbf{R}_{i,m} - \mathbf{R}_{i,g}\|_F$ and $e_{\mathbf{t}} = \frac{1}{N}\sum_{i=1}^{N}\|\mathbf{t}_{i,m} - \mathbf{t}_{i,g}\|_2$, where $\{(\mathbf{R}_{i,m}, \mathbf{t}_{i,m})\}_{i=1}^{N}$ denotes rigid transformations estimated by multi-view registration approach. In all multi-view registration approaches, the NN search method based on *k-d* tree [30] was utilized to establish point correspondences. All multi-view registration approaches were implemented in Matlab on a desktop with four-core 3.6GHz processor and 8GB of memory.

*4.1. Validation*

To validate the proposed approach, it is compared with two versions of multi-view registration approaches based on multi-scale descriptors: multi-view registration approach based on original correspondence propagation with rude augmentation and multi-view registration approach based on original correspondence propagation with model augmentation, which are abbreviated as OCPRA and OCPMA, respectively. As the proposed approach is based on fast correspondence propagation with model augmentation, it is abbreviated as FCPMA. Multi-view registration results are reported in runtime and registration errors. Fig. 7 displays the number of pair-wise registrations and runtime required to achieve multi-view registration for each approaches. Table 1 illustrates registration errors of each approach tested on four datasets.

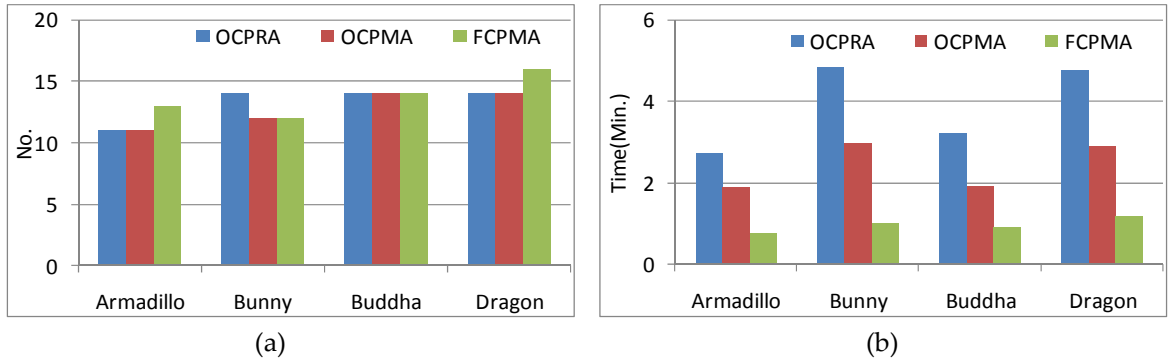

Fig. 7 Comparison of efficiency for different approaches based on multi-scale descriptors. (a) Required number of pair-wise registrations. (b) Run time.

As shown in Fig. 7, these three approaches require almost the same number of pair-wise registrations, which is a little more than the number of range scans to be aligned. As all these three approaches utilize the proposed reliability judgment, it is reasonable to conclude that this judgment is very effective for pair-wise registration. Sometimes, FCPMA may require a little more number of pair-wise registrations than other two approaches. In some scan pairs with low overlap percentages, there are a small number of reliable seed matches, which are all outside the top $30\%$. For these scan pair, the application of OCPRA and OCPMA may obtain good pair-wise registration results due to the application of correspondence propagation to all seed matches. Since FCPMA only applies the correspondence propagation to the top $30\%$ sorted seed matches, it is unable to get good registration results. Besides, OCPRA also may require a little more number of pair-wise registrations than other two approaches. This is because the rude augmentation results in many redundant points with multi-scale descriptors, which may reduce the robustness of pair-wise registration. Although FCPMA may require a little more number of pair-wise registrations, it is more efficient than other two approaches. By selecting the top $30\%$ sorted seed matches, it can accelerate the correspondence propagation, which will reduce the runtime of pair-wise registration. With the model augmentation, it deletes repeated range point with multi-scale descriptors so as to further improve the efficiency of subsequent registration. Therefore, both fast correspondence propagation and model augmentation are good for the efficiency of multi-view registration.

Table 1 Comparison of accuracy for different approaches based on multi-scale descriptors, where small value indicates good performance and bold number denotes the best result..

| Datasets | Scan No. | OCPRA | | OCPMA | | FCPMA | |
|---|---|---|---|---|---|---|---|
| | | $e_R$ | $e_t$ | $e_R$ | $e_t$ | $e_R$ | $e_t$ |
| Armadillo | 12 | **0.0068** | **0.5911** | 0.0099 | 0.7373 | 0.0096 | 0.7478 |
| Bunny | 10 | **0.0065** | **0.3356** | 0.0087 | 0.4877 | **0.0065** | 0.3615 |
| Buddha | 15 | 0.0230 | **1.3961** | 0.0242 | 1.4553 | **0.0225** | 1.4977 |
| Dragon | 15 | **0.0176** | **1.6343** | 0.0192 | 1.7520 | 0.0190 | 1.7384 |

As displayed in Table 1, all these three approaches can obtain accurate registration results and results of OCPRA are a little more accurate than that of other two approaches. This is because the rude augmentation keeps many redundant range points, which is good for accurate registration. However, it leads to high computation complexity. Since many exiting approaches can refine initial registration parameters, efficiency is more important than accuracy in global multi-view registration. Accordingly, OCPRA is not a good choice. For efficient registration, the proposed model augmentation is necessary to delete redundant range point with multi-scale descriptors. As most reliable seed matches contains small distances, it is easy to find good match set by the application of correspondence propagation to the top 30% sorted seed matches. Compared with raw correspondence propagation, fast correspondence propagation not only improves efficiency, but also obtains more accurate multi-view registration results.

In one word, the proposed approach is reasonable and effective for the global registration of unordered range scans.

*4.2. Comparsion*

To demonstrate its performance, the proposed approach was compared with two related approaches; there are automatic multi-view registration approach based on surface consistency (SurfC) [26] and automatic multi-view registration approach based on spanning tree (SpanT) [28]. Multi-view registration results are also reported in runtime and registration errors.

*4. 2.1 Accuacry and efficiency*

For comparison of efficiency and accuracy, experiments were carried on four data sets. Fig. 8 illustrates the number of pair-wise registrations and runtime required to achieve multi-view registration for all competed approaches. What's more, Table 2 displays registration errors of all competed approaches tested on four datasets. To demonstrate comparison in a more intuitive manner, Fig. 9 displays the registration results of four data sets for all competed approaches in the form of cross-sections.

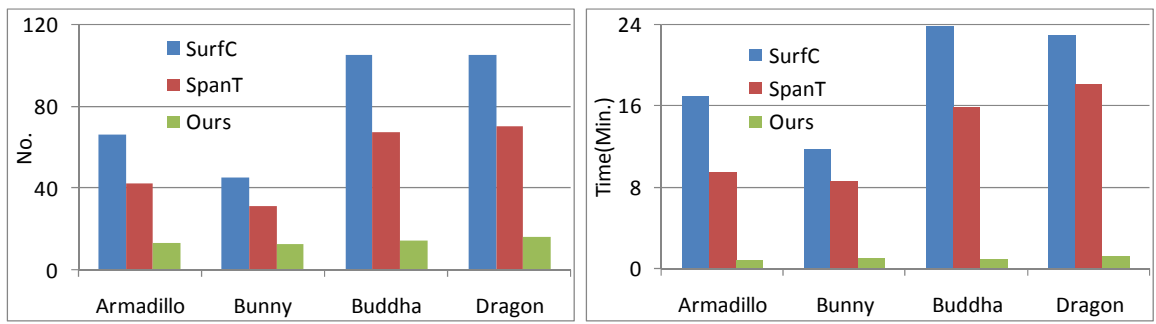

Fig. 8 Efficiency comparison of different approaches tested on four data sets. (a) Required number of pair-wise registrations. (b) Run time.

Table 2 Comparison of accuracy for different approaches, where small value indicates good performance and bold number denotes the best result.

| Datasets | Scan No. | SurfC [26] | | SpanT [28] | | Ours | |
|---|---|---|---|---|---|---|---|
| | | $e_R$ | $e_t$ | $e_R$ | $e_t$ | $e_R$ | $e_t$ |
| Armadillo | 12 | 0.0195 | 0.8467 | 0.0170 | 1.0233 | **0.0096** | **0.7478** |
| Bunny | 10 | 0.0262 | 1.4734 | 0.0242 | 1.3588 | **0.0065** | **0.3615** |
| Buddha | 15 | 0.0831 | 1.4952 | 0.0851 | 1.2455 | **0.0225** | **1.4977** |
| Dragon | 15 | 0.0245 | 1.4285 | 0.0219 | 1.7874 | **0.0190** | **1.7384** |

As shown in Fig. 8, SurfC requires more number of pair-wise registrations than other two approaches and the proposed approach requires the least number of pair-wise registrations among these competed approaches. Actually, SurfC should apply pair-wise registration to all scan pairs involved in multi-view registration. Given $N$ range scans, SurfC requires $N(N-1)/2$ pair-wise registrations. Therefore, this approach is inefficient. For multi-view registration, SpanT should find at least $(N-1)$ reliable pair-wise registration of raw scan pairs to construct completed spanning tree. As most of raw scan pairs contain low overlap percentages, it is also requires a large number of pair-wise registrations. While, the proposed approach only requires a small number of pair-wise registrations to achieve multi-view registration. During multi-view registration, the proposed approach utilizes the reliable pair-wise registration to augment the model shape, which can increase the overlap percentage between the model shape and data shape. With the increased overlap percentages, it become easy to obtain reliable pair-wise registration. Hence, the number of required pair-wise registrations is reasonably reduced. Since the proposed approach utilizes the pair-wise registration by fast correspondence propagation of multi-scale descriptors, it is more efficient than other two approaches.

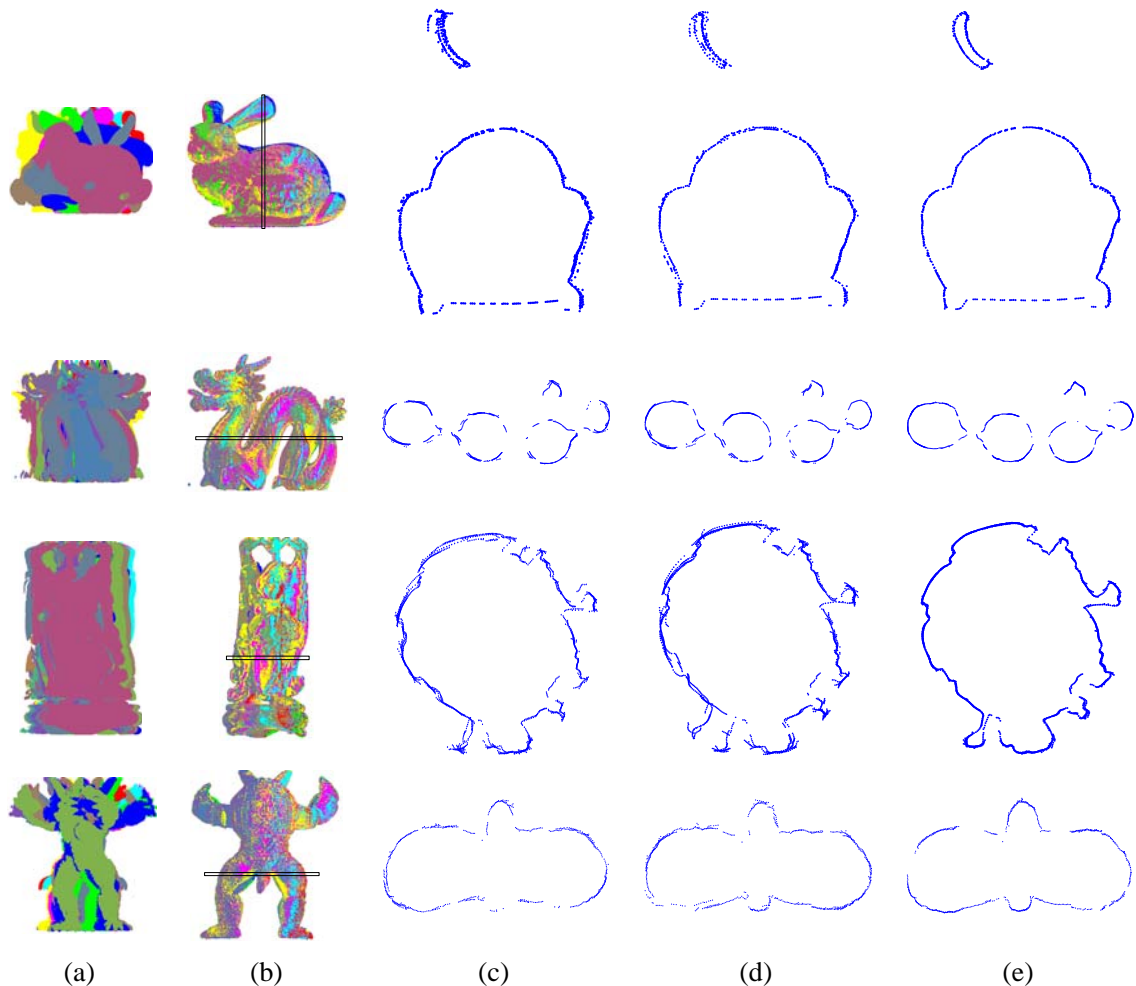

(a)　　　　　(b)　　　　　(c)　　　　　(d)　　　　　(e)

Fig. 9 Registration results of different approaches in the form of cross-section. (a) Unordered scans (b) Reconstructed 3D models. (c) Results of SurfC. (d) Results of SpanT. (e) Ours results.

As shown in Table 2 and Fig. 9, the proposed approach can obtain the most accurate results for multi-view registration among all competed approaches. Since these competed approaches utilize the pair-wise registration to achieve multi-view registration, they all suffer from the problem of error accumulation. In SurfC and SpanT, the pair-wise registration algorithm is directly applied to raw scan pairs. Since few raw scan pairs have high overlap percentages, few reliable results of pair-wise registration are very accurate, which will lead to large accumulated errors in multi-view registration. While, the proposed approach use reliable pair-wise registration results to augment model shape, which can increase overlap percentages of scan pairs to be aligned. With increased overlap percentages, results of pair-wise registration become more reliable and accurate, which will reduce accumulated errors in multi-view registration. Therefore, the proposed approach can always obtain the most accurate results for multi-view registration.

*4. 2.1 Robustness*

For the comparison of robustness, all competed approaches were tested on Stanford Bunny with five groups of different order. Multi-view registration results are also reported in runtime and registration errors Table 3 illustrated registration results of different approaches.

As shown in Table 3, all competed approaches are robust to the order of multiple range scans and the proposed approach can obtain the most accurate and efficient registration results for Stanford Bunny under different orders. Although the order of input range scans may be changed, all these three approaches utilize reliable pair-wise registration results to estimate rigid transformations for multi-view registration. Therefore, they can always achieve good results for multi-view registration. As SpanT and the proposed approach only apply the pair-wise registration on some scan pairs, the change of scan order will cause their required number of pair-wise registration to be varied. Accordingly, the run time of these two approaches is varied under different scan orders. However, SurfC applies pair-wise registration on all scans pairs involved in multi-view registration; its run time is relatively stable under different scan orders. By only considering the accuracy, the proposed approach is very robust.

Table 3 Comparison of robustness for different approaches, where small value indicates good performance and bold number denotes the best result.

| ID | SurfC [26] | | | SpanT [28] | | | Ours | | |
|---|---|---|---|---|---|---|---|---|---|
| | $e_R$ | $e_t$ | T(min.) | $e_R$ | $e_t$ | T(min.) | $e_R$ | $e_t$ | T(min.) |
| Order1 | 0.0262 | 1.4734 | 11.8039 | 0.0242 | 1.3588 | 8.5779 | **0.0065** | **0.3615** | **1.0044** |
| Order2 | 0.0280 | 1.2832 | 11.7961 | 0.0295 | 1.3697 | 6.0479 | **0.0075** | **0.3264** | **0.9908** |
| Order3 | 0.0346 | 1.9781 | 11.4876 | 0.0308 | 1.6397 | 9.3768 | **0.0062** | **0.4645** | **1.2782** |
| Order4 | 0.0360 | 2.1528 | 12.0357 | 0.0282 | 1.5158 | 9.4409 | **0.0052** | **0.3472** | **1.0956** |
| Order5 | 0.0289 | 1.7500 | 11.9747 | 0.0296 | 1.3778 | 6.2444 | **0.0078** | **0.3908** | **1.1849** |

## 5. Conclusions

This paper proposes the global approach for multi-view registration of unordered range scans. This approach is based on the pair-wise registration, which is achieved by the fast correspondence propagation of multi-scale descriptors. For multi-view registration, the reliability judgment is designed to choose reliable pair-wise registrations, which are utilized to augment the model shape. By model augmentation, the overlap percentage of scan pair will be reasonable increased. With the increased overlap percentage, the pair-wise registration is able to obtain reliable accurate results, which will reduce the required number of pair-wise registration and accumulated errors for multi-view registration. Experimental results carried on public available data sets illustrated that the proposed approach can robustly achieve multi-view registration of unordered range scans with good accuracy and efficiency.

Although the proposed approach has good performance, it does not mean that this approach can solve all multi-view registration problems. Given a set of unordered scans, there may be no other scan that contains high overlap percentage to one scan. If this special scan is viewed as the reference scan, it is difficult to obtain reliable pair-wise registration for this scan, which will lead to the failure of multi-view registration. However, it should be noted that all existing approaches for multi-view registration proposed so far share this limitation as well. In the future, we will extend this approach to robot mapping, especially for large-scale environment mapping.

**Acknowledgments:** Supported by the National Natural Science Foundation of China under Grant nos. 61573273 and 91648121.

**Author Contributions:** Jihua Zhu advised the study and contributed to writing in all phases of the work; Siyu Xu performed the experiments; Zutao Jiang conceived and designed the experiments; and Jun Wang and Zhongyu Li revised the manuscript.

**Conflicts of Interest:** The authors declare no conflict of interest.